\documentclass[journal]{IEEEtran}
\usepackage{amsmath,amsthm,amssymb,amsfonts,mathrsfs}
\usepackage{algorithmicx,algorithm}
\usepackage{array}
\usepackage{textcomp}
\usepackage{stfloats}
\usepackage{url}
\usepackage{verbatim}
\usepackage{graphicx}
\usepackage{cite}
\usepackage{bm}
\usepackage{algpseudocode}
\usepackage{scalerel,stackengine}
\stackMath
\newcommand\reallywidehat[1]{%
	\savestack{\tmpbox}{\stretchto{%
			\scaleto{%
				\scalerel*[\widthof{\ensuremath{#1}}]{\kern.1pt\mathchar"0362\kern.1pt}%
				{\rule{0ex}{\textheight}}
			}{\textheight}%
		}{2.4ex}}%
	\stackon[-6.9pt]{#1}{\tmpbox}%
}
\usepackage{hyperref}

\usepackage{soul, color, xcolor}
\soulregister{\cite}7 
\soulregister{\citep}7 
\soulregister{\citet}7 
\soulregister{\ref}7 
\soulregister{\pageref}7 

\theoremstyle{definition}

\usepackage{makecell}
\usepackage{graphicx}
\usepackage{textcomp}
\usepackage{array} 
\usepackage{longtable}
\usepackage{booktabs}
\usepackage{float}
\usepackage{threeparttable}
\usepackage{booktabs} 
\usepackage{multirow}

\begin{document}

\title{Data-Driven Modeling and Control for Tethered Space Systems with Koopman-Informed Graphs}

\author{Ao Jin, 
	Yifeng Ma, 
	Panfeng Huang, 
	Fan Zhang
	
\thanks{Ao Jin, Panfeng Huang and Fan Zhang are with Research Center for Intelligent Robotics, School of Astronautics, Northwestern Polytechnical University, Xi’an 710072, China. (E-mail: jinao@mail.nwpu.edu.cn)}
\thanks{Yifeng Ma is with Department of Mechanical and Automation Engineering, The Chinese University Hong Kong, 999077, China.}
}

\maketitle

\begin{abstract}
Modeling tethered space systems is critical for advanced orbital operations. Flexible components such as tethers and space nets are integral to these systems but present significant control challenges due to their high dimensional, strongly coupled, and nonlinear dynamics. While data driven methods offer alternative modeling approaches, they frequently struggle with long term predictive stability and spatial generalization. To address this, we propose the Koopman Graph Dynamics (KGD) framework to learn the structural dynamics by integrating the global linear evolution of the Koopman operator with the local topological priors of Graph Neural Networks. Building upon this representation, we develop a KGD based Model Predictive Control strategy for tethered space systems. Subsequently, the ground experiments on flexible tether and space net demonstrate the high precision modeling capabilities of the proposed method. Crucially, the framework exhibits exceptional capacity for spatial transfer without retraining. Models trained exclusively on small configurations successfully predict and control significantly larger, unseen physical scales. Furthermore, the orbit simulations within a physics engine verify the effectiveness of the proposed approach for tethered space systems.

\end{abstract}


\section{Introduction}

As orbital activities rapidly expand, future space missions demand lightweight, fuel-efficient robotic systems capable of highly adaptable operations across various workspaces. By offering unparalleled advantages in dynamic reconfigurability, minimal propellant consumption, and mission versatility, tethered space system (TSS) have emerged as a highly attractive architectural paradigm \cite{Wen2008a}. Within this category, tethered space robot (TSR) and tethered space net robot (TSNR) present a solutions for on-orbit servicing and active debris removal, capitalizing on their compliant capture mechanisms and extended operational ranges \cite{Zhao2019}. Furthermore, tethered formation system (TFS) broadens this utility by facilitating distributed, multi-spacecraft topologies for Earth observation and deep-space exploration \cite{Zhang2022a}. Collectively, these systems represent foundational TSS architectures, serving as crucial platforms for future space exploration missions.

Despite their immense application potential, the modeling and control of TSS remain fundamentally challenging. Across TSR, TSNR, and TFS, these architectures inherently exhibit spatially continuous dynamics, severe underactuation, and rigid-flexible coupling. The presence of space tethers strictly invalidates simplistic rigid-body assumptions, introducing highly nonlinear dynamics induced by their large deformations, elasticity, and spatial oscillations. \cite{Yu2018}. For TSR, this complexity escalates during orbital maneuvers as the time-varying tether length strongly couples with spatial librations \cite{Huang2016}. Parallel challenges emerge in TSNR, where flexible net generates unpredictable contact-induced transients post-impact \cite{Zhao2019}. Similarly, in TFS, tether elasticity actively induces structural oscillations that severely complicate stable formation maintenance \cite{Zhang2022a}. Consequently, synthesizing robust controllers for these platforms requires a framework capable of explicitly resolving these nonlinear continuum effects, strong rigid-flexible couplings, and underactuated constraints.

To address these challenges, extensive research has explored various strategies, yet a fundamental trade-off persists between modeling fidelity and control tractability. High-fidelity analytical formulations accurately capture continuum behaviors but render advanced optimal controller design computationally prohibitive. Conversely, control-oriented simplistic models facilitate real-time synthesis but entirely neglect inherent flexibility characteristics, often causing severe performance degradation during actual deployment. Recently, data-driven techniques based on Koopman operator theory have emerged to bypass strict first-principles modeling by identifying linear embeddings of nonlinear systems. However, current learning paradigms predominantly target rigid-body dynamics. They lack the structural priors necessary to encode the spatially distributed topologies of flexible continuum systems. Consequently, a critical gap remains in formulating a unified representation that simultaneously resolves the spatial topological complexities and nonlinearities of TSS to enable fast, adaptive optimal control.

Accordingly, this paper proposes a unified data-driven control framework for TSS. First, a graph-based representation is established to abstract the complex flexible continuum into a standardized topological structure. This spatial abstraction effectively encodes the flexibility and intricate couplings inherent in the TSS. Building upon this topological foundation, the framework introduces a computationally efficient model predictive control (MPC) scheme formulated within a globally linearized Koopman space. By projecting the highly nonlinear dynamics into a lifted space and leveraging state condensing techniques, the framework enables real-time optimal control synthesis without sacrificing modeling fidelity. Finally, to guarantee sustained robust performance against dynamic uncertainties, an asynchronous dual-rate architecture is developed. This mechanism intelligently decouples high-frequency control execution from low-frequency online model adaptation. By synergizing these mechanisms, this integrated approach systematically resolves the long-standing trade-off between accurate continuum modeling and real-time control tractability. 

\textbf{\textit{Main Contributions of this work}}:
\begin{itemize}
	\item A unified spatial-temporal representation is established by combining graph-based topological abstraction with Koopman operator theory. 
	\item A highly efficient predictive control algorithm is formulated within the lifted observable space. By exploiting state condensing on the globally linearized dynamics, this scheme circumvents severe computational bottlenecks to enable real-time optimal control for TSS.
	\item An asynchronous dual-rate online adaptation architecture is designed. This mechanism decouples high-frequency control execution from low-frequency model updates to effectively mitigate dynamic uncertainties and unmodeled continuum transients.
	\item Comprehensive validations of the proposed framework are conducted through high-fidelity simulator and ground experiments.
\end{itemize}

The rest of this paper is organized as follows. Section \ref{sec_related_work} provides a comprehensive review of related work. Section \ref{sec_preliminarie} establishes the necessary mathematical preliminaries and formally states the problem formulation. Section \ref{sec_main_results} details the main results. Section \ref{sec_simulation} and Section \ref{sec_experiment} validate the performance of the proposed methodology through high-fidelity simulation results and rigorous ground experiments, respectively. Finally, Section \ref{sec_conclusion} concludes this paper.

\section{Related Work} \label{sec_related_work}
\subsection{Tethered Space System}
\begin{figure}[!t]
	\centering
	\includegraphics[width=19pc]{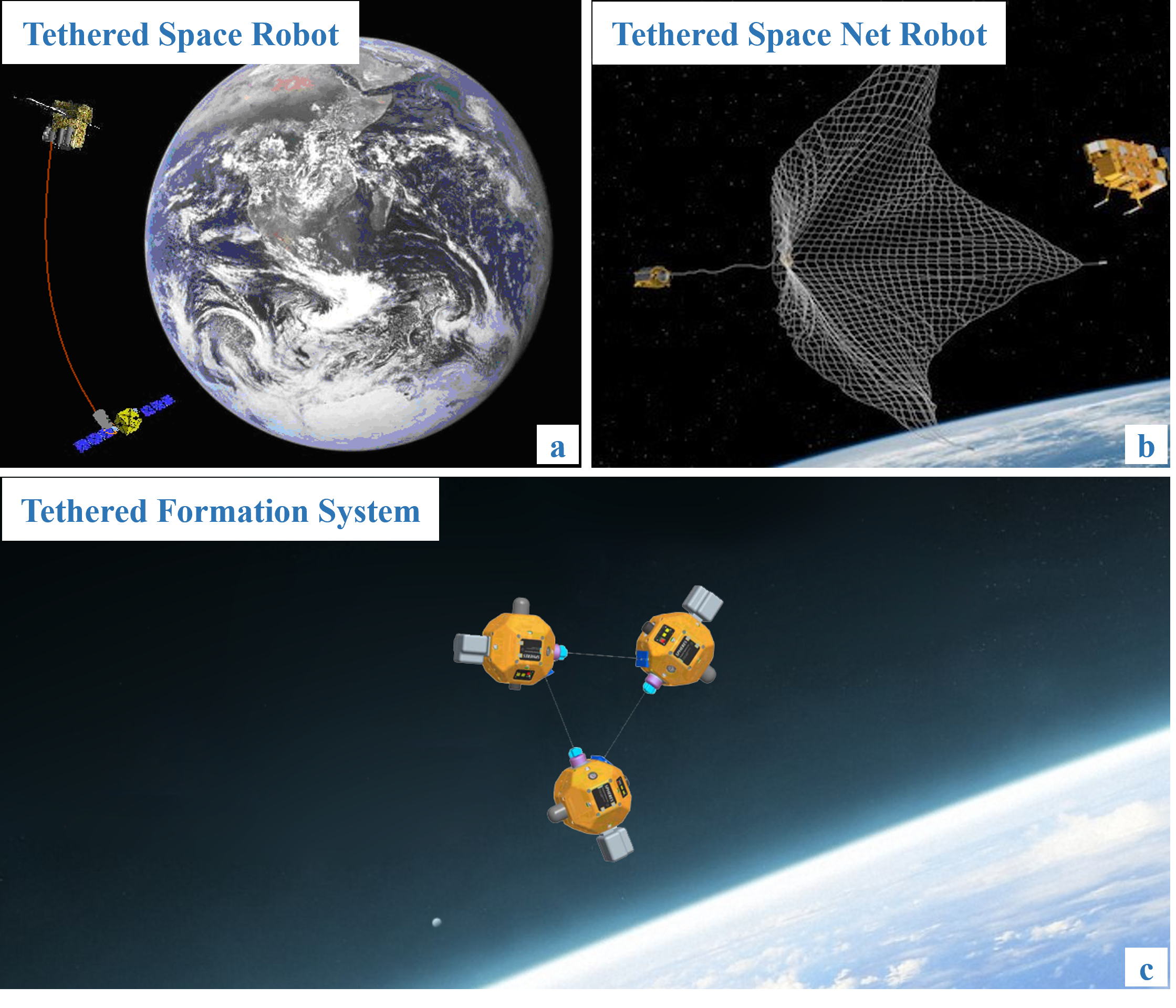} 
	\caption{TSS for diverse orbital applications: (a) TSR is composed of a space platform, a space tether, and a manipulator; (b) TSNR comprising four maneuvering units and a flexible net, which represents a promising solution for space debris removal; and (c) TFS orchestrated to maintain a precise and stable satellite geometric formation connected by space tethers.} \label{fig_tss_overview}
\end{figure}

The TSS discussed in this work includes the TSR, TSNR and TFS, which are shown in Fig. \ref{fig_tss_overview}. In this subsection, a brief introduction of TSS is given as follows. 

\textbf{Tethered Space Robot} Existing research on target capture and tether deployment maneuvers within TSR faces a fundamental trade-off between modeling fidelity and control tractability. To prioritize control tractability, control-oriented studies rely heavily on the simplistic dumbbell model \cite{Wen2008,Sun2014,Kang2019}, which facilitates straightforward controller synthesis by assuming a rigid, inelastic tether, yet entirely neglects the continuum elasticity and bending behaviors. Conversely, high-fidelity analytical models, such as the spatially discretized bead-rod \cite{Williams2008} and absolute nodal coordinate formulation (ANCF) \cite{Bai2023} models, accurately capture these complex flexible dynamics. However, their highly coupled and complicated nonlinear formulations render advanced controller design exceptionally challenging. Despite these analytical limitations, classical control methods (e.g., adaptive control and sliding mode control) based on simplified lumped-mass Lagrangian mechanics have achieved success in specific TSR maneuvers \cite{Huang2017}. More recently, to fundamentally bypass the intractability of high-fidelity models, the field is experiencing a progressive paradigm shift towards data-driven frameworks \cite{Jin2024}. 

\textbf{Tethered Space Net Robot} Compared with TSRs, the literature on TSNR has concentrated more specifically on contact dynamics and closing control during debris capture. A key challenge is that, after contact, the flexible net does not naturally wrap and close around the target, especially under off-center impact conditions. Accordingly, existing work has emphasized explicit modeling of post-contact net behavior and the associated risks of incomplete enclosure and tether damage \cite{Zhao2019}. Consequently, the current TSNR research focus centers on target enclosure, robustness against contact uncertainties, and vibration attenuation during the capture transient \cite{Zhao2020}. Despite these advances, deriving real-time, global optimal control methods for such high-dimensional system remains computationally prohibitive under traditional model-based paradigms.

\textbf{Tethered Formation System} Research on TFS fundamentally investigates multi-spacecraft configurations coupled by flexible links. Early literature primarily formulated the highly coupled attitude-orbit dynamics. To ensure formation maintenance, system stability was conventionally analyzed via reduced-order modeling and synchronization theory \cite{Chung2008}. More recently, studies have explicitly incorporated tether elasticity to address complex deployment maneuvers. This progression highlights a clear paradigm shift from simplified formation regulation towards high-fidelity modeling and deployment control design \cite{Zhang2022a}.

Crucially, elevating modeling fidelity across diverse physical geometries from TSR to TSNR and distributed TFS exposes a fundamental mathematical bottleneck. The literature currently lacks a unified, computationally tractable representation to systematically manage their complex topologies and high-dimensional nonlinear dynamics.

\subsection{Learning Dynamics for Prediction and Control}

Within the broader context of learning dynamics, recent research is shifting away from purely black-box identification. Instead, the focus has pivoted toward control-oriented latent representations that preserve analytical structures. Foundational data-driven approaches prioritize discovering parsimonious governing laws directly from data \cite{Brunton2022a}. This effectively bypasses the complexities of strict first-principles modeling. Notably, Koopman operator theory offers a powerful paradigm in this domain. It seeks finite-dimensional linear representations of nonlinear dynamics within lifted observable spaces \cite{Brunton2016a}. Consequently, this global linearization enables the direct application of linear control techniques to complex nonlinear systems \cite{Abraham2019,Ren2023a}. Addressing continuous-time dynamics, neural ordinary differential equations (NODEs) provide a highly scalable learning framework \cite{Chen2018}. Recent literature has seamlessly embedded NODEs within MPC pipelines, progressively incorporating physical priors like Lyapunov constraints and port-Hamiltonian mechanics to strictly guarantee system safety and robustness \cite{Revay2024,Duong2024}. Within the broader domain of robotic control, this paradigm has catalyzed the development of neural and meta-adaptive architectures designed explicitly for disturbance rejection and out-of-distribution generalization \cite{Wei2025,Wang2025b,Romero2026}. This evolutionary trend is equally prominent in aerospace systems. Recent advancements apply data driven control to complex aerospace maneuvers like spacecraft rendezvous and tethered robotics \cite{Yao2023,Servadio2023,Mao2026}.

Collectively, these works signify a transition towards control-aware neural dynamics that integrate physical structure into the control loop. However, current learning paradigms predominantly address rigid-body systems and struggle with the spatially distributed complexities of flexible continuum structures. This reveals a critical gap regarding coupled rigid and flexible dynamics, requiring a control methodology to effectively regulate such highly complex space topologies.

\section{Preliminaries And Problem Statement} \label{sec_preliminarie}

\subsection{Graph Neural Network}

A graph can be formally defined as $G = (\mathcal{O}, \mathcal{R})$, comprising a set of nodes (vertices) $\mathcal{O}$ and a set of interconnected edges $\mathcal{R}$. Generally, the nodes $\mathcal{O}$ or edges $\mathcal{R}$ own node or edge attributes. These features are mapped via a node attribute function $\bm{f}_O : \mathcal{O} \to \mathcal{A}$ and an edge attribute function $\bm{f}_R : \mathcal{R} \to \mathcal{W}$, where $\mathcal{A}$ and $\mathcal{W}$ denote the respective continuous feature spaces.

Graph Neural Networks (GNNs) leverage deep learning architectures to process this non-Euclidean graph data, aiming to extract high-dimensional latent representations through neural networks \cite{Thomas2023}. The fundamental learning pipeline of a GNN layer operates on a spatial message-passing paradigm, recursively updating the hidden representation of each node by aggregating information propagated from its local topological neighborhood via the connecting edges.

\subsection{Koopman Operator Theory}

Consider a nonlinear control system in discrete time
\begin{equation}
	\bm{x}_{k+1} = \bm{f}(\bm{x}_k, \bm{u}_k)
	\label{eq_nonlinear_system}
\end{equation}
where $\bm{x}_k \in \mathbb{R}^n$ is the state and $\bm{u}_k \in \mathbb{R}^m$ is the control input. The Koopman operator theory provides a framework to represent this finite-dimensional nonlinear system as an infinite-dimensional linear system. By defining a set of vector-valued lifting functions (or observables) $\bm{\Phi}: \mathbb{R}^n \rightarrow \mathbb{R}^K$ (where $K \gg n$), the original state $\bm{x}_k$ is mapped into a higher-dimensional feature space $\bm{z}_k = \bm{\Phi}(\bm{x}_k)$. To accommodate the control input within the linear framework, we adopt the assumption that control input $\bm{u}$ does not evolve in higher-dimensional feature space and the lifting functions take the form
\begin{equation}
	\bm{g}(\bm{x},\bm{u}) =\bm{\Phi}(\bm{x}) + \bm{L} \bm{u}
\end{equation}
where $\bm{L}$ is a constant matrix. In this sense, the nonlinear dynamics (\ref{eq_nonlinear_system}) can be reformulated as 
\begin{equation}
	\begin{gathered}
		\bm{\mathcal{K}}\bm{g}(\bm{x}_k,\bm{u}_k)=\bm{g}(\bm{x}_{k+1},0)=\bm{\Phi}(\bm{x}_{k+1})\\
		=\begin{bmatrix}\bm{A}&\bm{B}\end{bmatrix}\begin{bmatrix}\bm{\Phi}\left(\bm{x}\right)\\\bm{u}\end{bmatrix}
		\end{gathered}
\end{equation}
Consequently, the nonlinear dynamics can be represented by a lifted linear system (LLS) in the higher-dimensional feature space
\begin{equation}
	\begin{cases}
		\bm{z}_{k+1} = \bm{A} \bm{z}_k + \bm{B} \bm{u}_k \\
		\bm{x}_k = \bm{C} \bm{z}_k
	\end{cases}
	\label{eq_lifted_linear_system}
\end{equation}
where $\bm{A} \in \mathbb{R}^{K \times K}$ and $\bm{B} \in \mathbb{R}^{K \times m}$ represent the finite-dimensional approximation of the Koopman operator, and $\bm{C} \in \mathbb{R}^{n \times K}$ is the projection matrix used to recover the original physical state from lifted space.

While the Koopman operator provides an exact linear representation of the nonlinear dynamics, its infinite-dimensional nature renders it computationally intractable for controller design. To bridge this gap, a finite-dimensional approximation is practically mandated. Consequently, data-driven algorithms, such as the extended dynamic mode decomposition (EDMD), are employed to approximate the system matrices $\bm{A}$ and $\bm{B}$ from snapshot data. Given a dataset $\mathcal{D}$ consisting of $T$ transition snapshots, we define the following data matrices
\begin{equation}
	\begin{aligned}
		\bm{Z}_{1:T-1} = [\bm{z}_1, \bm{z}_2, \dots, &\bm{z}_{T-1}],\ \bm{Z}_{2:T} = [\bm{z}_2, \bm{z}_3, \dots, \bm{z}_T] \\
		\bm{U}_{1:T-1} &= [\bm{u}_1, \bm{u}_2, \dots, \bm{u}_{T-1}]
	\end{aligned}
	\label{eq_data_matrices}
\end{equation}
The finite-dimensional Koopman matrices can be obtained by solving the following least-squares minimization problem:
\begin{equation}
	[\bm{A}, \bm{B}] = \arg\min_{\bm{A}, \bm{B}} \left\| \bm{Z}_{2:T} - \begin{bmatrix} \bm{A} & \bm{B} \end{bmatrix} \begin{bmatrix} \bm{Z}_{1:T-1} \\ \bm{U}_{1:T-1} \end{bmatrix} \right\|_F^2
	\label{eq_edmd_optimization}
\end{equation}

\textit{Problem Statement:}
The central problem in governing TSS lies in resolving the fundamental contradiction between high-fidelity modeling and tractable controller design. In this paper, the primary objective is to design a unified mathematical framework capable of simultaneously achieving accurate spatial-temporal dynamic representation and efficient control synthesis to overcome this inherent trade-off.

\section{Main Results}\label{sec_main_results}
This section delineates the proposed data-driven control framework. We first establish a unified graph-based representation for TSS dynamics. Building upon this spatial abstraction, a fast model predictive control scheme is formulated via state condensing within the Koopman space. Finally, an asynchronous dual-rate architecture is introduced to enable online model adaptation for the robust performance against dynamic uncertainties.

\subsection{A Graph-Based Representation of Tethered Space System for Model-Based Control}
Across all configurations of TSS, including TSR, TSNR, and TFS, the highly flexible tethers and nets fundamentally dictate the system's dynamic behavior. Therefore accurately capturing these complex and high-dimensional dynamics is crucial. Leveraging GNNs to learn the true underlying dynamics in a purely data-driven manner is a promising approach. To bridge the physical continuum with the graph-based learning architecture, we spatially discretize the continuous flexible components into a finite set of interconnected point masses (i.e., discrete nodes). This unified topological abstraction allows us to establish a generalized modeling framework regardless of the specific TSS configuration. The proposed formulation is detailed as follows.

Following this spatial discretization, the generalized nonlinear dynamics of the entire TSS can be governed by a unified continuous-time differential equation:
\begin{equation}
	\dot{\bm{x}} = \bm{f}(\bm{x}, \bm{u})
	\label{eq:nonlinear_dynamics}
\end{equation}
where $\bm{x}$ denotes the aggregated global physical state of all discrete nodes, and $\bm{u}$ represents the control input applied to the system.

\begin{figure}[!t]
	\centering
	\includegraphics[width=21pc]{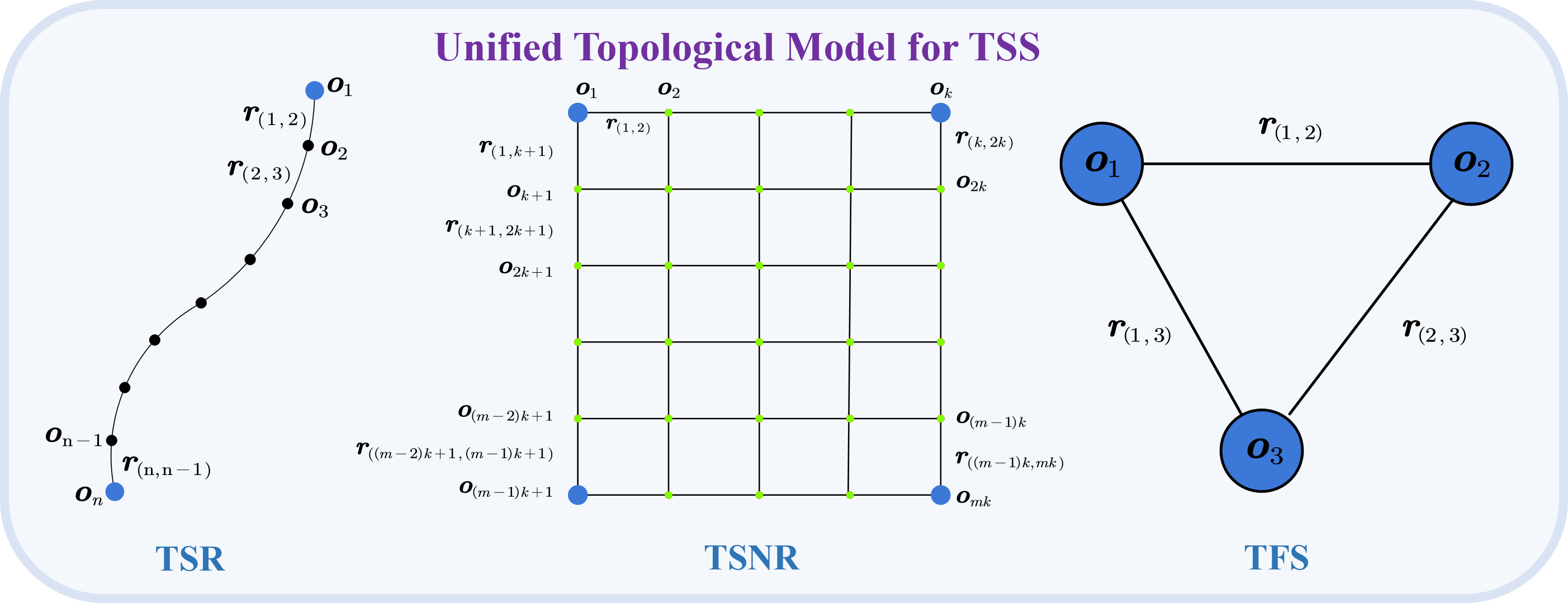} 
	\caption{Unified topological abstraction mapping physically diverse TSS into a generalized graph representation $G = (\mathcal{O}^t, \mathcal{R})$. \textit{(Left)} The TSR is discretized into a 1D chain topology. \textit{(Middle)} The TSNR is modeled as a 2D flexible mesh grid, governed by four maneuvering units. \textit{(Right)} The TFS is represented as a distributed multi-agent formation graph. Across all configurations, the large blue nodes ($o_i$) denote active control units, the small nodes denote passive lumped masses of the tether/net, and the connecting edges $\bm{r}_{(i,j)}$ denote the physical interactions between adjacent nodes.} \label{fig_tsr_discrete_nodes}
\end{figure}

As depicted in Fig. \ref{fig_tsr_discrete_nodes}, all instances within the TSS, whether they are the platform, the active maneuvering units, or the passive tether/net segments, are uniformly treated as interacting nodes in a topological space.  This physical configuration maps to a directed graph $G = (\mathcal{O}^t, \mathcal{R})$, where the set of vertices $\mathcal{O}^t := \{\bm{o}_i^t\}_{i=1}^N$ corresponds to these physical nodes at a specific discrete time step $t$. The node attribute at time $t$ is defined as a tuple $\bm{o}_i^t = (\bm{x}_i^t, \bm{u}_i^t, \bm{a}_i^o)$. Here, $\bm{x}_i^t \in \mathbb{R}^{n_x}$ is the state of node $i$ at time $t$, whereas $\bm{a}_i^o$ is a time-invariant one-hot vector differentiating the specific node types. Since the physical topology of every instance within the TSS remains constant during the operation, the set of edges $\mathcal{R} := \{\bm{r}_k\}_{k=1}^{N_e}$ is time-independent and represents the mechanical pair-wise interactions between the nodes. The edge attribute is defined as $\bm{r}_k = (m_k, n_k, \bm{a}_k^r)$, where $m_k$ and $n_k$ ($1 \leq m_k, n_k \leq N$) specify the indices of the connected nodes, and $\bm{a}_k^r$ denotes the specific relation type.

Following standard message-passing schemes \cite{Gilmer2017}, the spatial-temporal evolution of the graph is governed by an edge attribute function $\bm{f}_R$ and a node attribute function $\bm{f}_O$. The discrete-time dynamics update can be formulated as
\begin{equation}
	\begin{aligned}
		\bm{e}_k^t &= \bm{f}_R(\bm{o}_{m_k}^t, \bm{o}_{n_k}^t, \bm{r}_k), \quad k = 1, 2, \cdots, N_e \\
		\bm{o}_i^{t+1} &= \bm{f}_O \left( \bm{o}_i^t, \sum_{k \in \mathcal{S}_i} \bm{e}_k^t \right), \quad i = 1, 2, \cdots, N
	\end{aligned}
	\label{eq_gnn_update}
\end{equation}
where $\bm{e}_k^t$ represents the latent physical interaction force (message) passed along edge $k$, and the set $\mathcal{S}_i$ contains the indices of all edges pointing toward node $i$. 

While (\ref{eq_gnn_update}) effectively captures the propagation of tension waves along the tether/net, it inherently yields a black-box transition model. More critically, the control inputs $\bm{u}^t$ applied to the active nodes are deeply coupled within the implicit, nonlinear functions $\bm{f}_O$ and $\bm{f}_R$, which poses significant difficulties for controller design. To overcome this critical limitation, we integrate Koopman operator theory into the message-passing process, leading to our proposed \textbf{Koopman Graph Dynamics (KGD)}. Rather than relying on implicit black-box propagation, KGD employs a GNN as a trainable lifting function to map the highly coupled, nonlinear nodal states into a higher-dimensional feature space. This transformation allows us to construct a globally linear state-space representation directly from input/output data for TSS. Consequently, the explicit linear dynamics for any individual node $i$ can be mathematically represented as \footnote{As the one-hot vector $\bm{a}_i^o$ is static by definition, the first parameter $\bm{o}_i^t$ in $\bm{f}_O$ is replaced with $\bm{x}_i^t$.}
\begin{equation}
	\begin{cases}
		 \bm{e}_k^t = \bm{f}_R(\bm{o}_{m_k},\bm{o}_{n_k},\bm{r}_k) \\
		 \bm{g}_i^{t} = \bm{f}_O(\bm{x}_i^t,\sum_{k\in \mathcal{S}_i}\bm{e}_k^t) \\
		 \bm{z}_i^{t+1} = \bm{A}_i\bm{z}_i^t+\bm{B}_i\bm{u}_i^t + \bm{r}_i^t \\
		 \bm{x}_i^{t} = \bm{C}_i\bm{z}_i^t
	\end{cases} \label{KGD}
\end{equation}
where $\bm{z}_i^t = \bm{\Psi}_i(\bm x_i^t) = \bm{g}_i^{t}\in\mathbb{R}^K$ represents the lifting function associated with Koopman operator for node $i$, $\bm{u}_i^t\in \mathbb{R}^{n_u}$ denotes the control input for node $i$, $\bm{A}_i\in\mathbb{R}^{K\times K}$, $\bm{B}_i\in\mathbb{R}^{K\times n_u}$, $\bm{C}_i\in\mathbb{R}^{n_x\times K}$, $\bm{r}_i^t$ represents residual error of modeling for node $i$, $k=1... N^2$, $i=1... N$.

By aggregating the local nodal representations, the KGD for the entire TSS can be formulated in a compositional manner
\begin{equation}
	\begin{cases}
		\bm{z}^{t+1} = \bm{A} \bm{z}^t + \bm{B} \bm{u}^t + \bm{r}^t \\
		\bm{x}^t = \bm{C} \bm{z}^t
	\end{cases}
	\label{eq_compositional_kgd}
\end{equation}
where the global physical state $\bm{x}^t$, control input $\bm{u}^t$, and residual error $\bm{r}^t$ are obtained by concatenating their respective local nodal vectors
\begin{equation}
	\begin{aligned}
		\bm{x}^t &= [(\bm{x}_1^t)^\top, (\bm{x}_2^t)^\top, \cdots, (\bm{x}_N^t)^\top]^\top \\
		\bm{u}^t &= [(\bm{u}_1^t)^\top, (\bm{u}_2^t)^\top, \cdots, (\bm{u}_N^t)^\top]^\top \\
		\bm{r}^t &= [(\bm{r}_1^t)^\top, (\bm{r}_2^t)^\top, \cdots, (\bm{r}_N^t)^\top]^\top
	\end{aligned}
\end{equation}
Similarly, the global latent state $\bm{z}^t \in \mathbb{R}^{NK}$ is constructed through a compositional lifting function $\bm{\Phi}(\bm{x}^t)$, which stacks the individual local lifting functions $\bm{\Psi}_i(\bm{x}_i^t)$
\begin{equation}
	\bm{z}^t = \begin{bmatrix} \bm{z}_1^t \\ \bm{z}_2^t \\ \vdots \\ \bm{z}_N^t \end{bmatrix} = \begin{bmatrix} \bm{\Psi}_1(\bm{x}_1^t) \\ \bm{\Psi}_2(\bm{x}_2^t) \\ \vdots \\ \bm{\Psi}_N(\bm{x}_N^t) \end{bmatrix} \triangleq \bm{\Phi}(\bm{x}^t)
	\label{eq_global_lifting}
\end{equation}
A profound computational advantage of this compositional KGD framework is the decoupled structure of the global linear evolution matrices $\bm{A} \in \mathbb{R}^{NK \times NK}$ and $\bm{B} \in \mathbb{R}^{NK \times Nn_u}$. They naturally exhibit a strict block-diagonal topology
\begin{equation}
	\bm{A} = \begin{bmatrix} \bm{A}_1 & \bm{0} & \bm{0} \\ \bm{0} & \ddots & \bm{0} \\ \bm{0} & \bm{0} & \bm{A}_N \end{bmatrix}, \quad
	\bm{B} = \begin{bmatrix} \bm{B}_1 & \bm{0} & \bm{0} \\ \bm{0} & \ddots & \bm{0} \\ \bm{0} & \bm{0} & \bm{B}_N \end{bmatrix}
	\label{eq_block_diagonal}
\end{equation}

By actively exploiting this block-diagonal property, the structural complexity of the system is radically reduced. For a fully dense representation, the matrices $\bm{A}$ and $\bm{B}$ would contain $N^2K^2$ and $N^2Kn_u$ coupled elements, respectively. However, under our decoupled compositional KGD framework, the number of unknown parameters to be identified sharply shrinks to merely $NK^2$ for matrix $\bm{A}$, and $NKn_u$ for matrix $\bm{B}$. Utilizing the EDMD formulation (\ref{eq_edmd_optimization}), these decoupled parameter blocks can be highly efficiently estimated from input/output data, which is detailed as follows.

Consider a dataset $\mathcal{D}$ collected from the system trajectories, consisting of $T$ sequential data
\begin{equation}
	\mathcal{D}_{1:T} = [(\bm{x}^1, \bm{u}^1), \cdots, (\bm{x}^{T-1}, \bm{u}^{T-1}), \bm{x}^T]
	\label{eq_dataset}
\end{equation}
By employing the GNN as the parameterized lifting function $\bm{\Phi}$, the states $\bm{x}^t$ ($t = 1, 2, \dots, T$) are projected into the latent Koopman space. These lifted features are then grouped into corresponding snapshot data matrices
\begin{equation}
	\begin{aligned}
		\bm{Z}^{1:T-1} = [\bm{z}^1, \bm{z}^2, \cdots, \bm{z}^{T-1}]\ \bm{Z}^{2:T} = [\bm{z}^2, \bm{z}^3, \cdots, \bm{z}^T]
	\end{aligned}
	\label{eq_snapshot_matrices}
\end{equation}
Then matrices $\bm{A}$ and $\bm{B}$ can then be estimated by minimizing the one-step forward prediction error over the entire dataset. This translates to the following least-squares optimization problem
\begin{equation}
	\bm{A}, \bm{B} = \arg\min_{\bm{A}, \bm{B}} \left\| \bm{Z}^{2:T} - (\bm{A}\bm{Z}^{1:T-1} + \bm{B}\bm{U}^{1:T-1}) \right\|_F^2
	\label{eq_edmd_estimation}
\end{equation}
where $\bm{U}^{1:T-1} = [\bm{u}^1, \bm{u}^2, \cdots, \bm{u}^{T-1}]$.

As the lifting function $\bm{\Phi}$ is parameterized by the GNN, its network weights must be strictly optimized to guarantee the existence of an invariant and dynamically consistent linear subspace. However, in standard deep Koopman learning, this training process typically needs a state reconstruction loss to ensure that the physical states can be recovered from the latent space. Therefore, our proposed KGD architecture employs a state-inclusive embedding strategy (where the original state $\bm{x}$ is explicitly concatenated into the lifted feature $\bm{z}_i = [\bm{x}_i^\top, \bm{\Psi}(\bm{x}_i)^\top]^\top$), the projection matrix $\bm{C}_i = [\bm{I}, \bm{0}]$ is inherently predefined. This structural design analytically guarantees zero reconstruction error, eliminating the need for a reconstruction loss term. Consequently, our composite loss function focuses purely on dynamic consistency and comprises two critical predictive components.

The first component is the {multi-step forward prediction loss}. Rather than merely penalizing one-step transition errors, this term forces the KGD to yield accurate long-horizon predictions. Given the initial latent state $\bm{z}^1$ and a sequence of control inputs, the $j$-th step forward prediction $\hat{\bm{z}}^j$ ($j = 2, 3, \cdots, T$) is derived by recursively applying the nominal form of (\ref{eq_compositional_kgd}) where the residual term $\bm{r}^t$ is neglected
\begin{equation}
	\hat{\bm{z}}^j = \bm{A}^{j-1}\bm{z}^1 + \sum_{k=0}^{j-2} \bm{A}^k \bm{B} \bm{u}^{j-k-1}
	\label{eq_forward_rollout}
\end{equation}
where $\bm{A}^{j-1}$ represents the matrix power. The forward prediction loss is then defined as the mean squared error over the prediction horizon
\begin{equation}
	\mathcal{L}_{\text{forward}} = \frac{1}{T-1} \sum_{j=2}^{T} \left\| \bm{z}^j - \hat{\bm{z}}^j \right\|_2^2
	\label{eq_loss_fwd}
\end{equation}
By accumulating the errors over multiple steps, $\mathcal{L}_{\text{forward}}$ actively suppresses the compounding accumulation of errors during long-horizon propagation. The second component is the {multi-step backward prediction loss}. Enforcing time-reversibility significantly improves the spectral properties and stability of the identified Koopman operator. Starting from the terminal state $\bm{z}^{T}$, the backward predicted state at step $T-j$ is calculated using the inverse dynamics
\begin{equation}
	\hat{\bm{z}}^{T-j} = \bm{A}^{-j}\bm{z}^T - \sum_{k=1}^{j} \bm{A}^{-(j-k+1)} \bm{B} \bm{u}^{T-k}
	\label{eq_backward_rollout}
\end{equation}
where $\bm{A}^{-1}$ denotes the pseudo-inverse of the matrix $\bm{A}$. Correspondingly, the backward prediction loss evaluates the dynamic discrepancy in the reverse time direction
\begin{equation}
	\mathcal{L}_{\text{backward}} = \frac{1}{T-1} \sum_{j=1}^{T-1} \left\| \bm{z}^{T-j} - \hat{\bm{z}}^{T-j} \right\|_2^2
	\label{eq_loss_bwd}
\end{equation}
The total training objective is formulated as a weighted sum of the bidirectional predictions
\begin{equation}
	\mathcal{L}_{\text{total}} = \mathcal{L}_{\text{forward}} + \lambda \mathcal{L}_{\text{backward}}
	\label{eq_loss_total}
\end{equation}
where $\lambda$ is a hyperparameter balancing the forward and backward dynamic consistencies. By jointly minimizing $\mathcal{L}_{\text{total}}$, the resulting LLS strictly adheres to the true unknown dynamics in both time directions.

As reflected above, the state-inclusive architecture yields two fundamental advantages for both system identification and downstream control. First, it completely eliminates the need for an auxiliary decoder network and the associated reconstruction penalty. Since the physical state $\bm{x}^t$ can be exactly recovered from the latent state $\bm{z}^t$ through a trivial linear projection. This allows the optimization process to focus entirely on the multi-step linear prediction accuracy of the Koopman operator, without grappling with complex reconstruction trade-offs. Second, and most importantly for the subsequent MPC formulation, this explicit state preservation allows the cost function to be evaluated directly on the true physical coordinates. Because the spatial variables $\bm{x}^t$ are an explicit subset of the augmented state $\bm{z}^t$, their corresponding tracking errors can be directly and independently penalized within the MPC cost matrix. This enables the solver to explicitly minimize the true physical tracking objective, while simultaneously maintaining appropriate regularizations on the latent dynamics $\bm{\Psi}(\bm{x}^t)$. This explicitly anchored formulation mathematically circumvents the decoding distortions inherent in traditional latent-space MPC.

\subsection{Fast MPC via State Condensing in Koopman Space}
Controlling TSR, TSNR, and TFS is fundamentally challenged by their highly underactuated continuum dynamics, where local inputs at the extremities propagate through flexible links to cause severe time delays and nonlinear couplings. Thus, the control framework must actively predict flexible deformations while satisfying strict operational constraints, such as actuator saturation and safe tension limits. To address this challenge, we propose a novel data-driven control framework termed \textbf{KGD-MPC}. By seamlessly embedding our globally linear KGD representation into the predictive formulation, KGD-MPC transforms the inherently intractable, non-convex continuum control problem into a convex Quadratic Program (QP). Nevertheless, solving this QP directly in the high-dimensional Koopman latent space remains computationally demanding. If the full state trajectory and control sequence are optimized simultaneously, the total number of decision variables expands to $N_p(NK + n_u)$, where $N_p$ is the prediction horizon. Since standard interior-point QP solvers scale cubically with the number of decision variables, this yields a prohibitive optimization complexity of $\mathcal{O}((N_p(NK + n_u))^3)$. To achieve real-time closed-loop execution, we formulate a fast predictive control scheme utilizing a dense state condensing technique. By leveraging the linear structure of KGD, we analytically eliminate the intermediate state variables $\bm{z}^t$ over the entire prediction horizon. This mathematical transformation maps the large sparse QP into a significantly smaller dense QP. This dramatically reduces the dimensionality of the decision space, enabling extremely efficient real-time execution.

Given the initial global latent state $\bm{z}^t$ at time step $t$, the nominal future states over the prediction horizon $N_p$ can be recursively derived from the nominal form of (\ref{eq_compositional_kgd})
\begin{equation}
	\begin{aligned}
		\bm{z}^{t+1} &= \bm{A}\bm{z}^t + \bm{B}\bm{u}^t \\
		\bm{z}^{t+2} &= \bm{A}^2\bm{z}^t + \bm{A}\bm{B}\bm{u}^t + \bm{B}\bm{u}^{t+1} \\
		&\vdots \\
		\bm{z}^{t+N_p} &= \bm{A}^{N_p}\bm{z}^t + \sum_{i=0}^{N_p-1} \bm{A}^{N_p-1-i}\bm{B}\bm{u}^{t+i}
	\end{aligned}
	\label{eq_recursive_expansion}
\end{equation}
To construct the dense QP formulation, we group the predicted state trajectory and the control sequence into augmented vectors $\bm{Z} \in \mathbb{R}^{N_p n_Z}$ and $\bm{U} \in \mathbb{R}^{N_p n_U}$, respectively
\begin{equation}
	\bm{Z} = \begin{bmatrix} \bm{z}^{t+1} \\ \bm{z}^{t+2} \\ \vdots \\ \bm{z}^{t+N_p} \end{bmatrix}, \quad
	\bm{U} = \begin{bmatrix} \bm{u}^t \\ \bm{u}^{t+1} \\ \vdots \\ \bm{u}^{t+N_p-1} \end{bmatrix}
	\label{eq_augmented_vectors}
\end{equation}
where $n_Z=NK$ and $n_U=Nn_u$ denote the dimensions of the global latent state and control input, respectively. The entire nominal state trajectory over the horizon can then be compactly rewritten in a fully vectorized form
\begin{equation}
	\bm{Z} = \bm{\mathcal{M}}\bm{z}^t + \bm{\mathcal{C}}_u\bm{U}
	\label{eq_vectorized_trajectory}
\end{equation}
where the matrices $\bm{\mathcal{M}}$ and $\bm{\mathcal{C}}_u$ are strictly defined by
\begin{equation}
	\begin{aligned}
		\bm{\mathcal{M}} &= \begin{bmatrix} \bm{A}^\top & (\bm{A}^2)^\top & \cdots & (\bm{A}^{N_p})^\top \end{bmatrix}^\top \\
		\bm{\mathcal{C}}_u &= \begin{bmatrix} 
			\bm{B} & \bm{0} & \cdots & \bm{0} \\ 
			\bm{A}\bm{B} & \bm{B} & \cdots & \bm{0} \\ 
			\vdots & \vdots & \ddots & \vdots \\ 
			\bm{A}^{N_p-1}\bm{B} & \bm{A}^{N_p-2}\bm{B} & \cdots & \bm{B} 
		\end{bmatrix}
	\end{aligned}
	\label{eq_condensing_matrices}
\end{equation}

The objective of the MPC is to track a desired latent reference trajectory $\bm{Z}_{\text{ref}}$, which yields a standard quadratic cost
\begin{equation}
	J = \frac{1}{2} (\bm{Z} - \bm{Z}_{\text{ref}})^\top \bm{\mathcal{Q}} (\bm{Z} - \bm{Z}_{\text{ref}}) + \frac{1}{2} \bm{U}^\top \bm{\mathcal{R}} \bm{U}
	\label{eq_cost_function}
\end{equation}
where $\bm{\mathcal{Q}}$ and $\bm{\mathcal{R}}$ are positive-definite block-diagonal weighting matrices. By substituting the condensed state equation into the cost function, we isolate the control sequence $\bm{U}$ as the sole decision variable
\begin{equation}
	J = \frac{1}{2} \bm{U}^\top \bm{\mathcal{H}} \bm{U} + \bm{f}^\top \bm{U}
	\label{eq_condensed_cost}
\end{equation}
where the condensed Hessian matrix $\bm{\mathcal{H}} \in \mathbb{R}^{N_p n_U \times N_p n_U}$ and $\bm{f} \in \mathbb{R}^{N n_U}$ are given by
\begin{equation}
	\begin{aligned}
		\bm{\mathcal{H}} &= \bm{\mathcal{C}}_u^T \bm{\mathcal{Q}} \bm{\mathcal{C}}_u + \bm{\mathcal{R}} \\
		\bm{f} &= \bm{\mathcal{C}}_u^T \bm{\mathcal{Q}} (\bm{\mathcal{M}} \bm{z}^t - \bm{Z}_{\text{ref}})
	\end{aligned}
	\label{eq_hessian_gradient}
\end{equation}

Because the matrices $\bm{A}$ and $\bm{B}$ remain constant over the short prediction horizon, the highly condensed Hessian matrix $\bm{\mathcal{H}}$ can be precomputed. Consequently, the MPC formulation is strictly reduced to a dense quadratic program
\begin{equation}
	\begin{aligned}
		\min_{\bm{U}} \quad & \frac{1}{2} \bm{U}^\top \bm{\mathcal{H}} \bm{U} + \bm{f}^\top \bm{U} \\
		\text{s.t.} \quad & \bm{U}_{\text{min}} \leq \bm{U} \leq \bm{U}_{\text{max}}
	\end{aligned}
	\label{eq_qp_formulation}
\end{equation}
This dimensionality reduction transforms the $\mathcal{O}((N_p(n_Z+n_U))^3)$ optimization complexity into $\mathcal{O}((N_p n_U)^3)$. Given that the control dimension $n_U$ is substantially smaller than the latent dimension $n_Z$, this mathematical substitution enables extremely efficient closed-loop execution.

\subsection{Online Adaptation via Asynchronous Dual-Rate Architecture}

Although the proposed KGD framework yields a globally linear representation trained on extensive offline datasets, the true continuum dynamics of TSS are inherently time-varying. In addition, during complex orbital operations, the system is continuously subjected to unmodeled external disturbances (e.g., orbital perturbations). Relying solely on a static, offline-identified Koopman model may inevitably lead to cumulative local approximation errors and degraded control performance over time. In this sense, it is necessary to continuously adapt the matrices $\bm{A}$ and $\bm{B}$ online using real-time input/output data. However, incorporating online system identification directly into the control loop introduces a severe computational burden. Updating the Koopman operator mandates solving a large-scale least-squares problem over a sliding window, followed by the recomputation of the dense condensed matrices ($\bm{H}$ and $\bm{f}$) for the QP solver. Executing these computationally intensive algebraic operations synchronously would severely block the high-frequency feedback control loop. 

To circumvent this computational bottleneck, we propose an asynchronous dual-rate control architecture. This framework physically decouples the real-time control loop and the online learning processes into two parallel execution threads operating at distinct frequencies. As delineated in Algorithm \ref{alg_dual_rate_mpc}, the \textit{High-Frequency Control Thread} dedicatedly samples the state, executes the lightweight GNN encoder, and solves the condensed dense QP problem at the control rate. Concurrently, the \textit{Low-Frequency Model Update Thread} operates asynchronously in the background. It collects recent state-action trajectories from a sliding window buffer to perform the least-squares updates. Once the new dynamics matrices ($\bm{A}_{\text{new}}, \bm{B}_{\text{new}}$) and their corresponding dense Hessian matrices are computed, they are seamlessly injected into the control thread. This dual-rate paradigm completely isolates the heavy computational burdens while endowing the proposed algorithm with online adaptability.

\begin{algorithm}[htbp]
	\caption{Asynchronous Dual-Rate Adaptive KGD-MPC}
	\label{alg_dual_rate_mpc}
	\begin{algorithmic}[1]
		\Require Initial system matrices $\bm{A}, \bm{B}$, reference trajectory $\bm{Z}_{\text{ref}}$, sliding window buffer $\mathcal{B}$, control period $\Delta t_\text{c}$, identification period $\Delta t_{\text{id}}$
		\State \textbf{Initialize:} Precompute condensing matrices $\bm{\mathcal{M}}, \bm{\mathcal{C}}_u$, and the dense Hessian $\bm{H}$ 
		\State \textbf{Spawn} Low-Frequency Shadow Thread
		\vspace{0.15cm}
		\hrule
		\vspace{0.15cm}
		\State \textbf{[Thread 1: High-Frequency Control Loop]} \Comment{Executes at $1/\Delta t_\text{c}$}
		\While{Control task is active}
		\State Sample current physical state $\bm{x}^t$
		\State Encode state-inclusive latent feature $\bm{z}^t$ 
		\State Fetch latest precomputed matrices $\bm{\mathcal{M}}, \bm{\mathcal{C}}_u$, $\bm{H}$, and $\bm{f}$
		\State Solve standard dense QP (\ref{eq_qp_formulation}) to obtain optimal control sequence $\bm{U}^*$
		\State Apply the first control input of $\bm{U}^*$ to the system
		\State Add data tuple $(\bm{z}^{t-1}, \bm{u}^{t-1}, \bm{z}^t)$ to $\mathcal{B}$
		\EndWhile
		\vspace{0.15cm}
		\hrule
		\vspace{0.15cm}
		\State \textbf{[Thread 2: Low-Frequency Model Update Thread]} \Comment{Executes at $1/\Delta t_{\text{id}}$}
		\While{Control task is active}
		\State Wait for update interval $\Delta t_{id}$
		\State Extract recent trajectory window $\mathcal{D} \leftarrow \mathcal{B}$
		\State Construct snapshot matrices $\bm{Z}^{1:T-1}, \bm{U}^{1:T-1}, \bm{Z}^{2:T}$ from $\mathcal{D}$
		\State Solve the optimization problem (\ref{eq_edmd_estimation}) for $\bm{A}_{\text{new}},\ \bm{B}_{\text{new}}$
		\State Rebuild condensed matrices $\bm{\mathcal{M}}_{\text{new}}, (\bm{\mathcal{C}}_u)_{\text{new}}$ using (\ref{eq_condensing_matrices})
		\State Precompute new dense Hessian using (\ref{eq_hessian_gradient})
		\State Update $\bm{\mathcal{M}} \leftarrow \bm{\mathcal{M}}_{\text{new}}, \bm{\mathcal{C}}_u \leftarrow (\bm{\mathcal{C}}_u)_{\text{new}}, \bm{H} \leftarrow \bm{H}_{\text{new}}$
		\EndWhile
	\end{algorithmic}
\end{algorithm}

\section{Ground Experimental Evaluation}\label{sec_experiment}
In the TSS, the highly nonlinear, strongly coupled, and high-dimensional dynamics of flexible tethers/nets make accurate explicit analytical modeling a formidable challenge. The proposed framework addresses this bottleneck by providing a purely data-driven modeling paradigm. Before extrapolating these models to orbital simulations, it is imperative to validate the framework's fundamental capability to capture real-world physical dynamics. Therefore, in this section, we conduct a comprehensive experimental evaluation. The primary objective is to verify that the proposed method can reconstruct the high-fidelity dynamics of one-dimensional tethers and two-dimensional nets directly from data. In addition, to intuitively demonstrate the practical utility and control authority of the learned models, a challenging active "shape control" task of a flexible tether is introduced. 

\subsection{Experimental Setup and Data Collection}
\begin{figure}[!t]
	\centering
	\includegraphics[width=20pc]{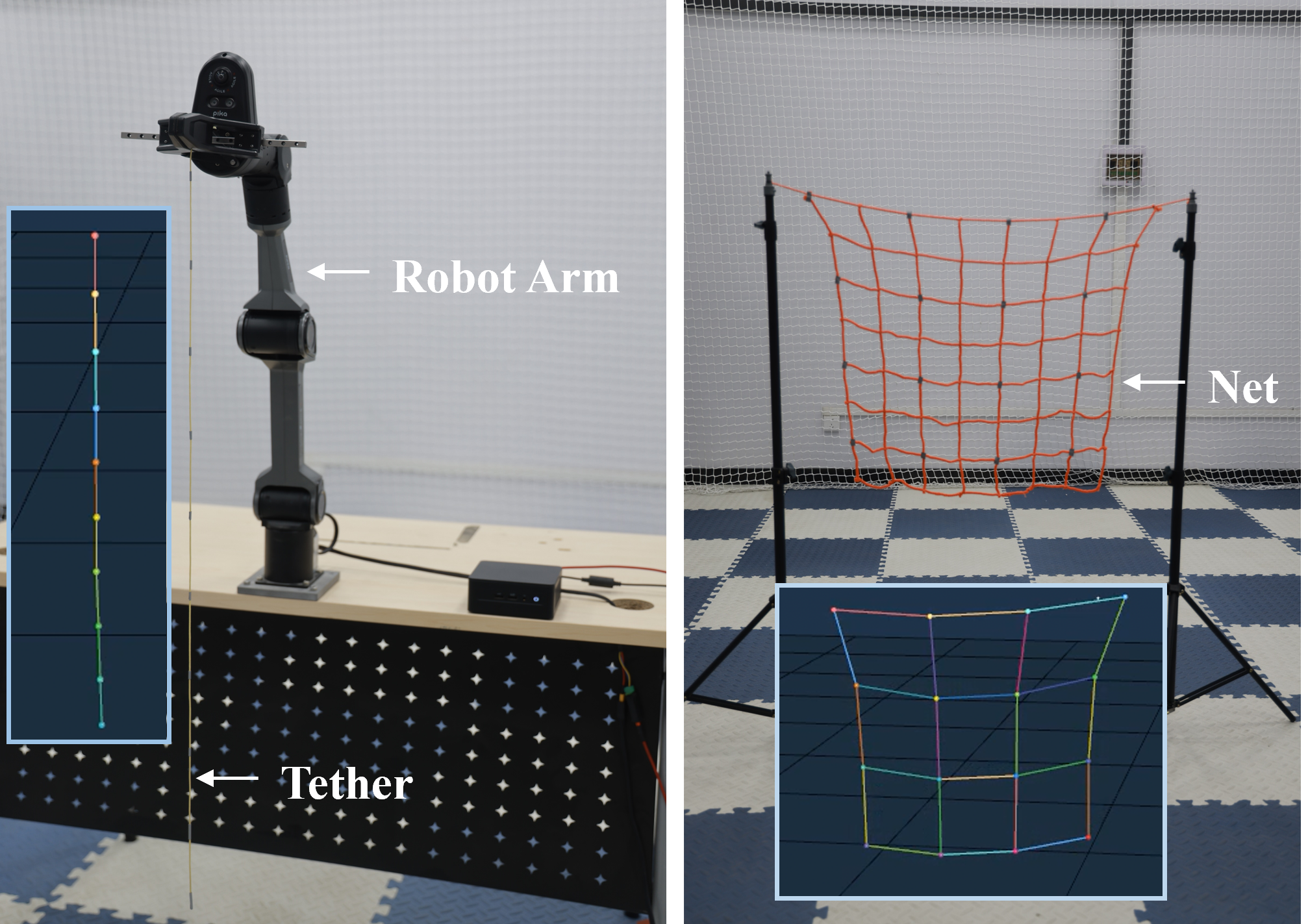} 
	\caption{The ground experimental setup for dynamics learning. The discretized node positions on both the tether and net are precisely tracked using a Nokov motion capture system, enabled by reflective markers affixed to each discretized node.} \label{fig_experiment_setup}
\end{figure}

\textbf{Scenarios}: 1) 1D Flexible Tether: This scenario is designed to model the fundamental dynamics of a flexible tether. The physical tether has a total length of 0.9 m and is uniformly discretized into 10 nodes, resulting in a spatial resolution of 0.1 m between adjacent nodes. One end of the tether is attached to a robot arm constrained to move along a single horizontal axis.  2) 2D Net: This scenario scales the validation to a more complex, strongly coupled flexible structure. The physical square net features a side length of 0.6 m and is discretized into a uniform grid with 4 nodes along each edge, yielding a constant spatial spacing of 0.2 m. One entire edge of the net is rigidly fixed to a support, while the opposite edge remains unconstrained and is free to fall under the gravity. Fig. \ref{fig_experiment_setup} presents the setup of ground experiments. 

\textbf{Data Collection}: Dynamic datasets for both scenarios were collected at a sampling interval of 0.01 s (100 Hz) by applying random kinematic excitations to their respective boundaries. For the 1D tether, randomly moving the robot arm along its axis generated 400 trajectories (100 steps each), partitioned into 320 for training and 80 for validation. For the 2D space net, data collection was conducted by rigidly fixing one entire edge while initializing the two corner nodes on the opposite free edge at random spatial positions. The subsequent motion under gravity yielded 600 trajectories (100 steps each), splitting into 480 for training and 120 for validation.

\textbf{Baseline Methods}: To comprehensively evaluate the proposed KGD, we compare it against three representative data-driven baselines: the MLP, which serves as an unstructured baseline mapping flattened inputs to future states to verify the necessity of explicit topological modeling; the Interaction Network (IN) \cite{Battaglia2016}, a structured physical reasoning model that explicitly separates relational and object reasoning to compute interaction effects; and the EDO-Net \cite{Longhini2023}, a state-of-the-art graph dynamics model that employs an adaptation module to extract latent physical properties from observation sequences and predicts future states via GNN. 

\textbf{Architecture of KGD}: The system state is first processed through three independent encoders for nodes, edges, and control inputs. Each encoder consists of a linear projection, followed by Layer Normalization and a SiLU activation function. These modules project the raw inputs into a unified 128-dimensional hidden space. The hidden features are then concatenated and processed by two consecutive layers. These layers utilize a multi-head attention mechanism (4 heads) while explicitly embedding spatial physical constraints to aggregate features and underlying mechanics. To capture global nonlinearities, the aggregated graph features are first lifted into a 256-dimensional Koopman latent space. Subsequently, a low-rank Koopman operator is applied. This operator is designed to evolve the high-dimensional latent states linearly. The evolved latent representations are mapped back to node-wise states by a two-layer MLP decoder. We used Adam with a learning rate of $10^{-3}$ and weight decay equal to $10^{-5}$.

\subsection{Evaluation of Learned Dynamics}
This subsection evaluates the KGD framework's predictive fidelity and scalability on the 1D flexible tether and 2D Net. 

\begin{table}[!t]
	\renewcommand{\arraystretch}{1.3}
	\caption{Comparison results of prediction errors (MSEs) of learned dynamics for tether}
	\label{table_kgd_evaluate_more_nodes}
	\centering
	\begin{tabular}{lccc}
		\toprule
		\textbf{Node Number} & \textbf{Method} & \textbf{$\bm E_{\text{0-150\ step}}$ [m]} & \textbf{$\bm E_{\text{150-30\ step}}$ [m]}   \\
		\midrule
		
		\multirow{4}{*}{n=12}
		& MLP & 3.430e+00 & 1.180e+01 \\
		& IN & 1.621e-03 & 3.932e-02 \\
		& EDO-Net & 4.019e-02 & 4.175e-01 \\
		& KGD & \textbf{1.013e-04} & \textbf{7.384e-03} \\
		
		\midrule
		
		\multirow{4}{*}{n=16}
		& MLP & 9.640e+00 & 4.170e+01 \\
		& IN & 1.063e-01 & 4.367e-01 \\
		& EDO-Net & 4.584e-02 & 2.461e-01 \\
		& KGD & \textbf{1.582e-04} & \textbf{1.355e-02} \\
		
		\midrule
		
		\multirow{4}{*}{n=20}
		& MLP & 1.287e+01 & 6.362e+01 \\
		& IN & 1.444e-01 & 4.705e-01 \\
		& EDO-Net & 5.448e-02 & 3.536e-01 \\
		& KGD & \textbf{8.326e-04} & \textbf{2.032e-02} \\
		
		\bottomrule
	\end{tabular}
\end{table}

To assess the generalization capabilities of KGD on 1D flexible tether, the model is verified on configurations with $n=12, 16, \text{and } 20$ nodes. In these cases, the nodal spacing is kept constant at 0.1 m, which results in tethers of increased total lengths. These tests examine the framework's ability to extend its learned local dynamics to larger physical scales while maintaining consistent modeling precision. Table \ref{table_kgd_evaluate_more_nodes} presents the quantitative comparison of prediction errors across different spatial discretizations and temporal horizons. As shown, the unstructured MLP baseline exhibits severe divergence across all scenarios, confirming its inability to generalize to unseen topological dimensions without explicit structural priors. While the graph-based baselines, IN and EDO-Net, demonstrate moderate short-term predictive capabilities, they suffer from substantial error accumulation over the extended temporal horizon (${150\  \text{step}-300\  \text{step}}$) and experience noticeable performance degradation as the system scale expands. Conversely, the proposed KGD framework consistently outperforms all baseline methods across every test configuration. Notably, even under the most challenging out-of-distribution scenario (n=20), KGD maintains a remarkably low short-term error of $8.326\times 10^{-4}$ m.

\begin{figure}[!t]
	\centering
	\includegraphics[width=21pc]{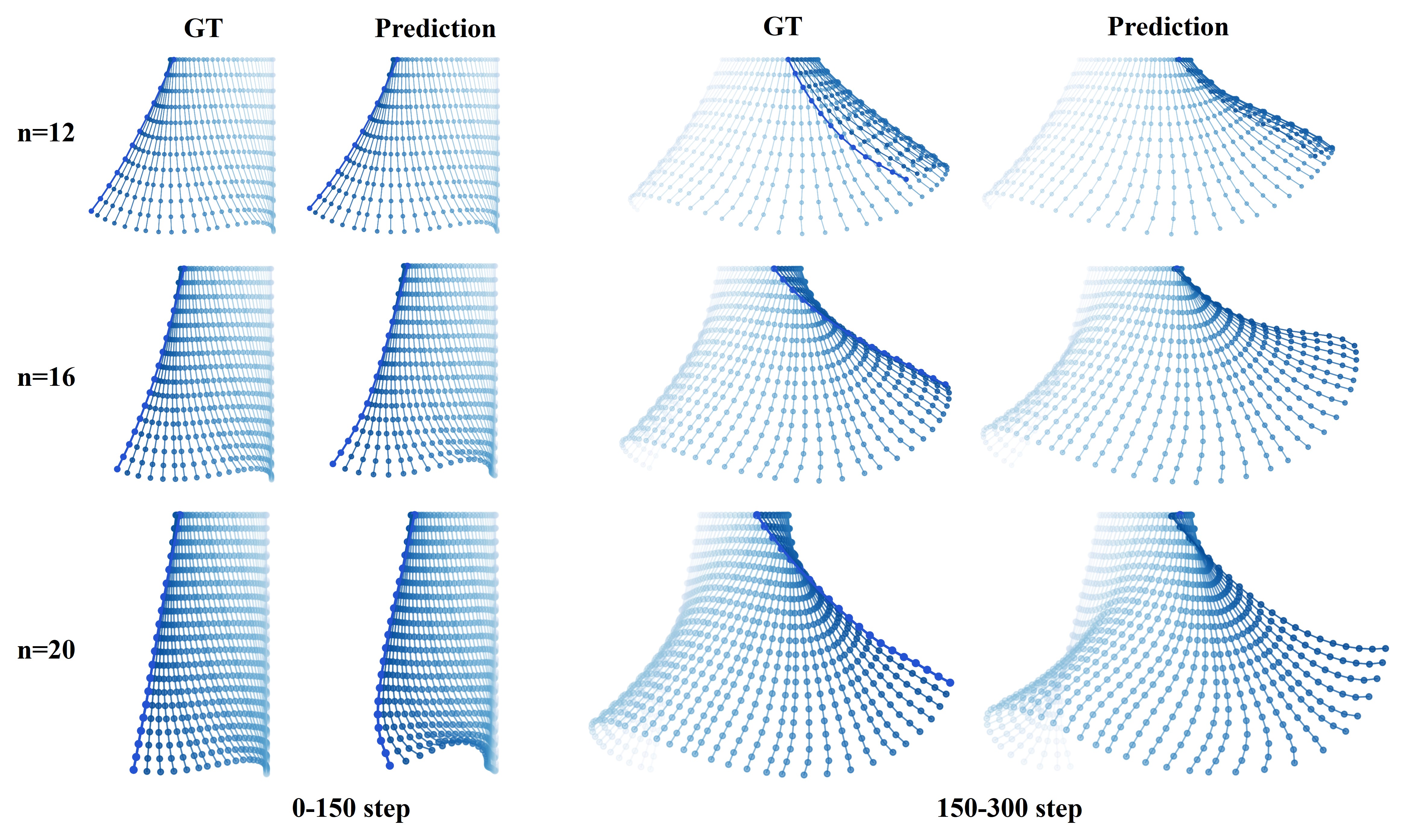} 
	\caption{Comparison of the trajectory rollouts for the 1D flexible tether. Ground-truth and predicted trajectories are compared across different spatial discretizations ($n=12, 16, 20$) and temporal horizons ($0-150 step$ s and $150-300 step$ s). The progression of time is visualized through structural states with a light-to-dark blue color gradient.} \label{fig_rope_evaluation}
\end{figure}

To further corroborate the quantitative findings, Fig. \ref{fig_rope_evaluation} provides a visual comparison between the ground-truth (GT) multi-step trajectories and the KGD predictions. The rollouts are depicted across three varying system scales ($n=12, 16, \text{and } 20$) and are separated into two distinct temporal horizons ($0-150$ step and $150-300$ step). As observed in the initial phase from 0 to 150 step, the KGD model accurately captures the transient structural deformation and whipping dynamics of the extended tethers. Crucially, during the extended horizon of 150 step to 300 step, baseline methods like MLP, IN, and EDO-Net typically suffer from severe error accumulation and structural distortion. In contrast, the proposed KGD continues to accurately capture the underlying dynamics and successfully reproduces the complex, high-dimensional nonlinear dynamics.

\begin{table}[!t]
	\renewcommand{\arraystretch}{1.3}
	\caption{Comparison results of prediction errors (MSEs) of learned dynamics for net} \label{table_net_kgd_more_nodes}
	\centering
	\begin{tabular}{lccc}
		\toprule
		\textbf{Grid Configuration} & \textbf{Method} & \textbf{$\bm E_{\text{0-500\ step}}$ [m]} & \textbf{$\bm E_{\text{500-1000\ step}}$ [m]}   \\
		\midrule
		
		\multirow{4}{*}{$6\times 6$}
		& MLP & 1.981e-01 & 4.568e-01    \\
		& IN & 1.965e-01 & 2.389e-01    \\
		& EDO-Net & 7.905e-02  & 3.597e-01   \\
		& KGD & \textbf{3.608e-02}  & \textbf{1.371e-01}    \\
		
		\midrule
		
		\multirow{4}{*}{$8\times 8$}
		& MLP & 4.397e-01 & 1.158e+00    \\
		& IN & 4.123e-01 & 7.822e-01    \\
		& EDO-Net & 3.439e-01  & 9.420e-01   \\
		& KGD & \textbf{2.753e-01}  & \textbf{4.803e-01}    \\
		
		\bottomrule
	\end{tabular}
\end{table}

Following the 1D flexible tether experiments, we further evaluate the predictive fidelity and spatial scalability of the KGD framework on a more complex, strongly coupled 2D net. The network trained on the $4 \times 4$ grid discretization is directly deployed to predict the dynamics of larger nets featuring $6 \times 6$ and $8 \times 8$ nodes with a uniform spatial spacing of 0.1 m. Table \ref{table_net_kgd_more_nodes} presents the comparison of prediction errors for the 2D net dynamics. Consistent with previous findings, the unstructured MLP baseline exhibits the highest errors across all scenarios, confirming its inability to capture the complex and coupled dynamics of the 2D net. While the structured baselines, IN and EDO-Net, show reasonable short-term predictive capabilities on the $6 \times 6$ grid, they suffer from significant error accumulation during the extended horizon from 500 step to 1000 step. Furthermore, as the system scale expands to the $8 \times 8$ configuration, the performance of these baseline methods deteriorates rapidly. In contrast, the proposed KGD framework consistently outperforms all baselines across both spatial discretizations and temporal horizons. Notably, during the challenging extended horizon (500 -1000 step) for the $8 \times 8$ net, KGD achieves an MSE of 4.803e-01 m, which substantially suppresses the catastrophic drift observed in other methods. 

\begin{figure}[!t]
	\centering
	\includegraphics[width=12pc]{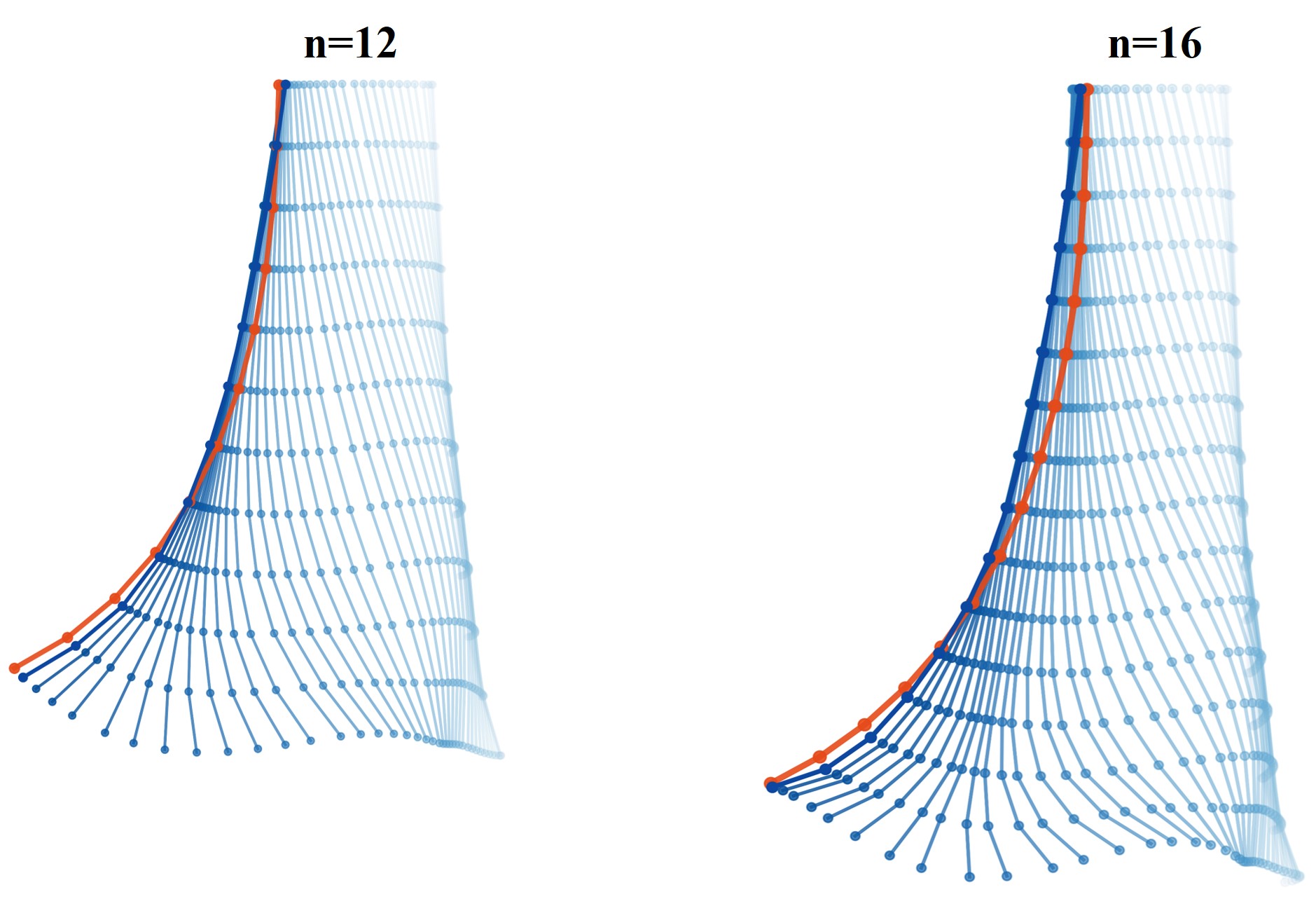} 
	\caption{The shape control execution is evaluated across unseen spatial discretizations ($n=12 \text{ and } 16$). The progression of time is visualized through structural states with a light-to-dark blue color gradient, while the solid orange line indicates the desired target shape.} \label{fig_shape_control}
\end{figure}

Collectively, these comprehensive evaluations across both the flexible tether and the space net conclusively demonstrate the superiority of the proposed KGD framework, highlighting its exceptional long term predictive stability and robust capacity for spatial generalization.

\subsection{Shape Control of Tether}
Building upon the high-fidelity dynamic predictions and spatial scalability established in the previous evaluation, this part extends the analysis to a downstream closed-loop application involving the active shape control of a flexible mass rope. To rigorously validate the effectiveness of our framework, we tackle the control of an $n=16$ tether utilizing the KGD model exclusively trained on the $n=10$ configuration. We employ the proposed KGD-MPC method (Algorithm \ref{alg_dual_rate_mpc}) to drive the underactuated flexible tether toward the target shape, where the velocity of robot arm is restricted within [-1, 1] m/s.  The following experiments aim to demonstrate that the superior spatial generalization capabilities of the KGD framework directly translate into effective, precise, and computationally viable physical control. The reference trajectory $\bm{Z}_{\text{ref}}$ is generated by mapping a predefined target shape into the 256-dimensional latent space through the trained lifting function $\bm{\Phi}$. The state weight matrix is set to $\bm{\mathcal{Q}} = 15\bm{I}_{256}$, where $\bm{I}_{256} \in \mathbb{R}^{256 \times 256}$ represents the identity matrix, and the control input weight matrix is configured as $\bm{\mathcal{R}} = 100I$. The control period is defined as $\Delta t_c = 0.02$ s. Concurrently, the matrices $\bm{A}$ and $\bm{B}$ update at an identification period of $\Delta t_{\text{id}} = 0.2$ s. The prediction horizon is set to $N_p=50$. The model updates are supported by a sliding window buffer $\mathcal{B}$ with a fixed length of 100.

The results of the KGD-MPC execution are illustrated in Fig. \ref{fig_shape_control}. As depicted by the blue sequential gradients, the flexible rope smoothly evolves from its initial state and accurately converges to the desired target shape. Notably, the KGD-MPC achieves precise shape tracking across both the $n=12$ and $n=16$ configurations. This successful zero-shot spatial transfer demonstrates that the KGD representation effectively captures the essential control-oriented dynamics, enabling robust and scalable closed-loop manipulation of complex flexible structures even outside the training distribution.

\section{Orbital Simulation}\label{sec_simulation}
\begin{figure}[!t]
	\centering
	\includegraphics[width=20pc]{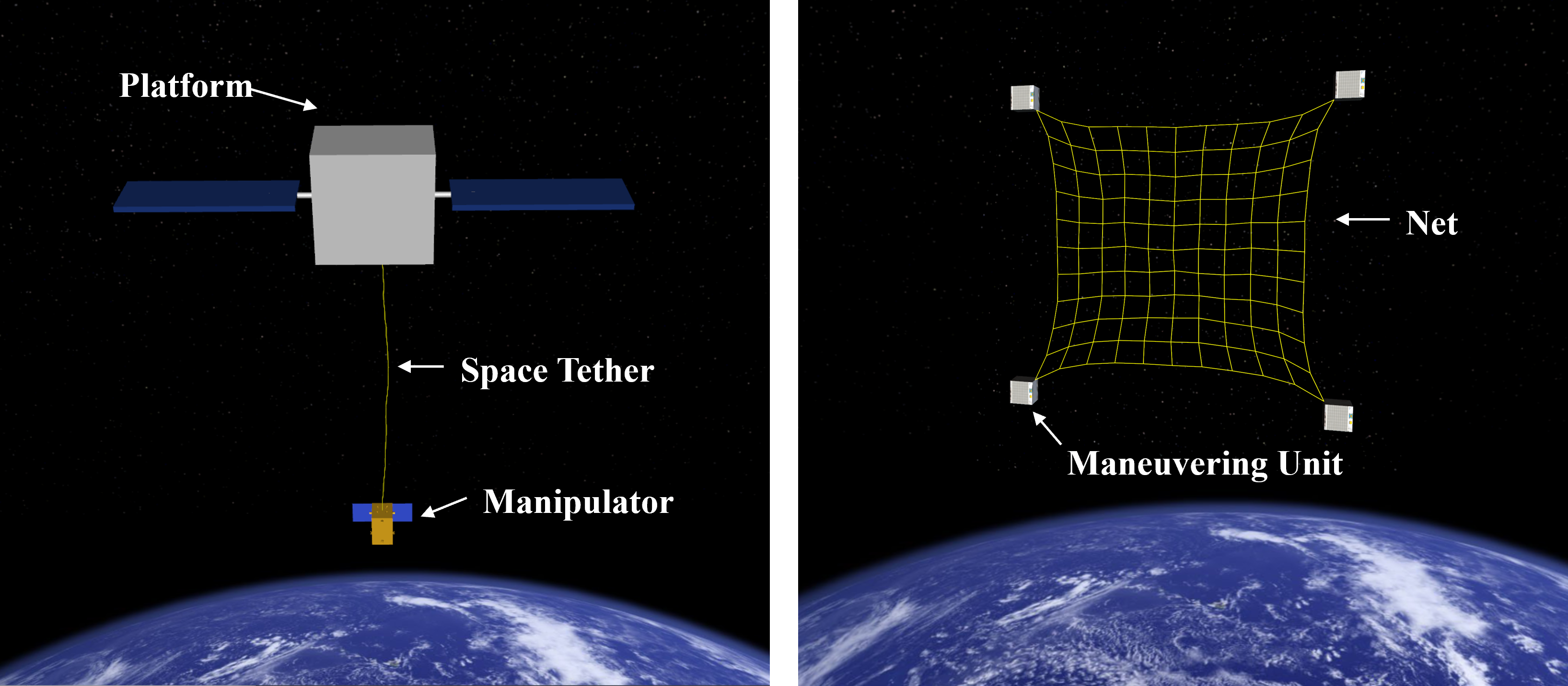} 
	\caption{High-fidelity simulation scenarios of TSR and TSNR in the MuJoCo physics engine.} \label{fig_simulator}
\end{figure}

This section extends the evaluation of proposed framework to a full-scale orbital environment. A high-fidelity simulator based on the MuJoCo physics engine \cite{Todorov2012} is developed, as shown in Fig. \ref{fig_simulator}. The primary objective is to rigorously verify the modeling and control effectiveness of the proposed KGD-MPC method on TSR and TSNR.

\subsection{Setup}
The parameters and initial conditions of the orbital simulation are listed as follows. For TSR, the length of tether $L=100$ m (is uniformly discretized into 101 nodes and regulated by a reel mechanism capable of applying a tension bounded within [0, 2] N), orbital velocity $\Omega = 0.00117\ \text{rad/s}$, mass of platform $m_{\text{p}}=2000$ kg, mass of manipulator $m_\text{m}=50$ kg, state weight matrix $\bm{\mathcal{Q}}_{\text{tsr}} = 10\bm{I}_{256}$, the control input matrix  $\bm{\mathcal{R}}_{\text{tsr}} = 50\bm{I}$, the control period $\Delta t_c^{\text{tsr}} = 0.02$ s, identification period $\Delta t_{\text{id}}^{\text{tsr}} = 0.2$ s, the initial velocity for deployment $\bm{v}_0=1.5$ m/s, and the control objective is to deploy the manipulator to the place 100 m away from where it starts deployment. For TSNR, the side length of the flexible net $L = 5$ m (is uniformly discretized into 11 nodes per edge with a spatial spacing of 0.5 m), mass of each maneuvering unit $m_u = 10$ kg (four units in total equipped with continuous three axis thrusters bounded within [-5, 5] N),  state weight matrix $\bm{\mathcal{Q}}_{\text{tsnr}} = \bm{I}_{256}$, control input matrix  $\bm{\mathcal{R}}_{\text{tsnr}} = 2\bm{I}$, the control period $\Delta t_c^{\text{tsnr}} = 0.05$ s, identification period $\Delta t_{\text{id}}^{\text{tsnr}} = 0.5$ s, and the objective is to capture a defunct satellite located at an initial distance of 40 m from the TSNR system. The size of sliding window buffer $\mathcal{B}$ is 200, and prediction horizon is set to $N_p=50$.

The physical properties of the flexible components, such as the stiffness and damping coefficients of the tether or space net, remain strictly unknown to the proposed algorithm. And for data collection, instead of simulating the full TSR/TSNR systems, the tether and the space net are isolated from the rigid body platforms and subjected to random excitation inputs directly within the physics engine. The dataset configuration for the orbital simulation, including the sampling interval and data allocation, remains identical to that of the previous ground experiments. The architecture of KGD is consistent with Section \ref{sec_experiment}.A.

\subsection{Evaluation of KGD-MPC}

\begin{figure}[!t]
	\centering
	\includegraphics[width=20pc]{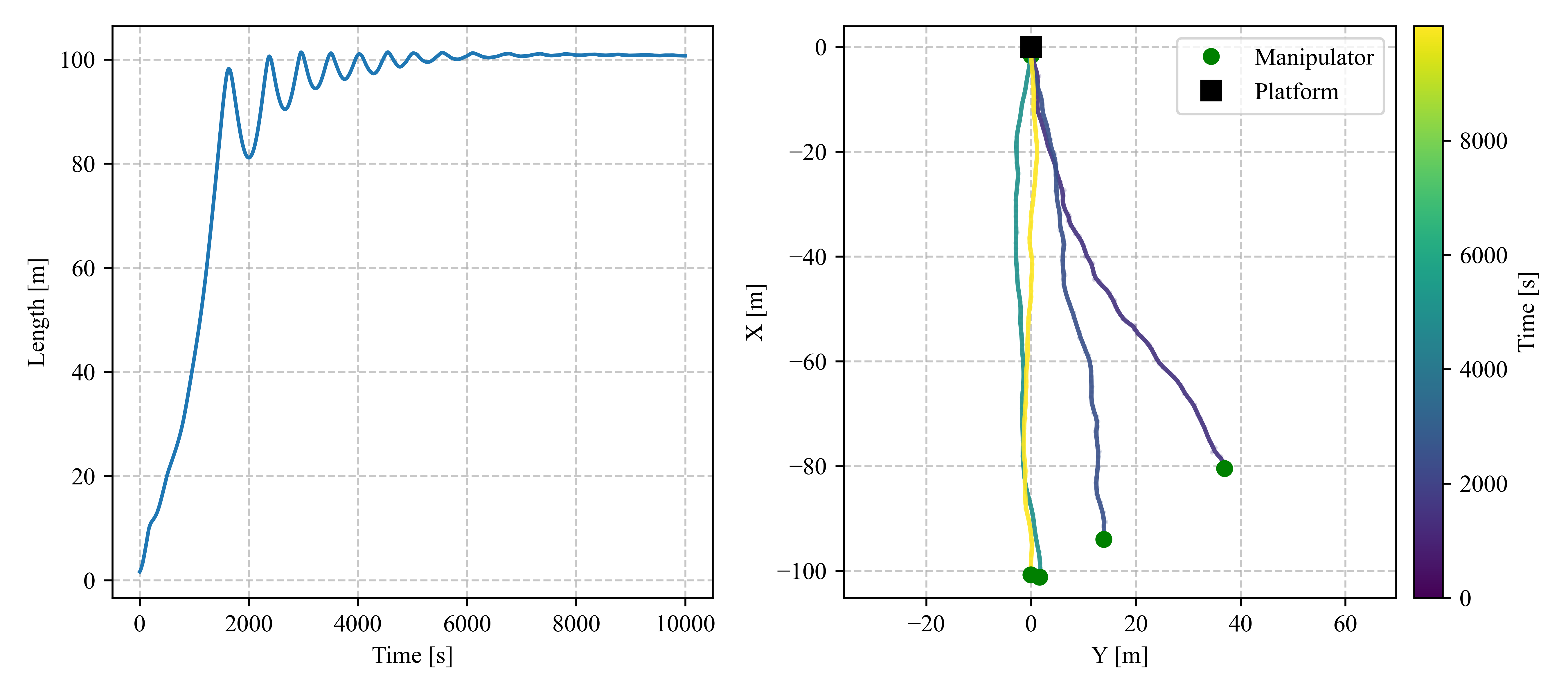} 
	\caption{Tether length and deployment trajectory for TSR.} \label{fig_orbital_sim_tsr}
\end{figure}

\begin{figure}[!t]
	\centering
	\includegraphics[width=20pc]{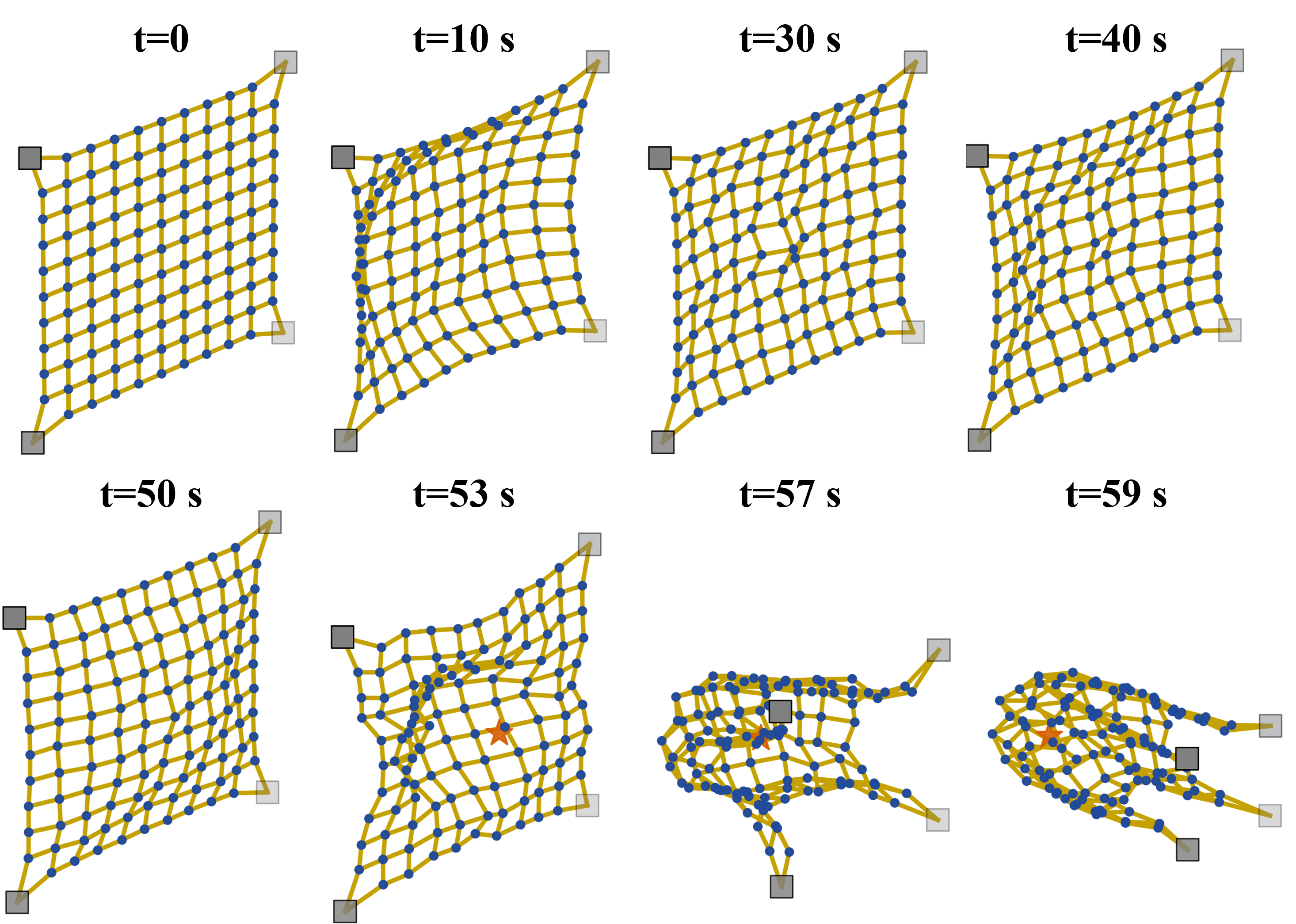} 
	\caption{Sequential snapshots of the defunct satellite capture for TSNR.} \label{fig_orbital_sim_tsnr}
\end{figure}

The results of KGD-MPC for TSR and TSNR are shown in Fig. \ref{fig_orbital_sim_tsr} and Fig. \ref{fig_orbital_sim_tsnr}. For TSR, the Algorithm \ref{alg_dual_rate_mpc} achieves precise tether length regulation at 100 m with slight oscillation while successfully deploying the manipulator to the target location. For TSNR, the proposed algorithm coordinates four maneuvering units to maintain the net shape during the approach phase and ultimately executes successful wrapping around the defunct satellite. Collectively, these complex orbital tasks validates the effectiveness and robustness of the proposed KGD MPC algorithm for the TSS.

\section{Conclusion} \label{sec_conclusion}
In this paper, we presented the Koopman Graph Dynamics (KGD) framework to address the complex modeling and control challenges inherent in the TSS. By integrating the Koopman operator with GNNs, the KGD approach enables high fidelity, data driven modeling of flexible tethers and nets purely from data. Extensive ground experiments and orbital simulations demonstrate that our method substantially mitigates long term error accumulation compared to unstructured baselines. Crucially, the framework exhibits an exceptional capability for spatial transfer without retraining, allowing models trained on small configurations to successfully predict the dynamics of significantly larger physical scales. Furthermore, the KGD can be updated online, enabling dynamic adaptation to disturbances. Building upon these predictive models, the KGD-MPC method successfully accomplishes the control task of TSS. In all, this paper provides a highly accurate, scalable, and computationally viable solution for the TSS. Future work will focus on the real orbital experimental validation of the proposed method.

\bibliographystyle{IEEEtran}
\bibliography{bibtex.bib}

\end{document}